\documentclass[pdflatex,sn-mathphys-num]{sn-jnl}

\usepackage{graphicx}%
\usepackage{multirow}%
\usepackage{amsmath,amssymb,amsfonts}%
\usepackage{amsthm}%
\usepackage{mathrsfs}%
\usepackage[title]{appendix}%
\usepackage{xcolor}%
\usepackage{textcomp}%
\usepackage{manyfoot}%
\usepackage{booktabs}%
\usepackage{algorithm}%
\usepackage{algorithmicx}%
\usepackage{algpseudocode}%
\usepackage{listings}%

\usepackage{fontawesome5}
\usepackage{etoolbox}

\theoremstyle{thmstyleone}%
\theoremstyle{thmstyletwo}%

\theoremstyle{thmstylethree}%
\newtheorem{definition}{Definition}%

\makeatletter
\def\blfootnotetext{\xdef\@thefnmark{}\@footnotetext}
\makeatother

\begin{document}

\title[ ]{Transformer autoencoder with local attention for sparse and irregular time series with application on risk estimation}

\author*{\fnm{Panteleimon} \sur{Rodis}
\href{https://orcid.org/0000-0001-9169-8202} 
\orcid{}}

\abstract{This paper introduces a framework specifically designed for sparse and irregular time series {risk estimation}. It is based on a Transformer Autoencoder with local attention, which leverages the powerful pattern identification capabilities of transformers complemented by traditional data cleaning and normalization methods. It efficiently captures relevant patterns within irregular sequences suffering from sparse data collection, benefiting from the discriminative ability of the local attention mechanism.
The proposed framework is applied to a real-world case study, on the risk estimation of non-technical losses in electrical power systems in a wide area in Greece.
Non-technical losses in electrical power systems, primarily stemming from electricity theft, pose significant economic and operational challenges. Detecting these anomalies is particularly challenging due to the inherent sparse and irregular nature of real-world data collection practices. Traditional risk estimation methods struggle with effectively capturing long-range dependencies and robustly handling such data characteristics.
We demonstrate that our approach effectively yields highly discriminative latent features, which results in more consistent risk estimation compared with existing state-of-the-art and widely used methods.  
It achieves high recall and precision, meeting the critical objectives of the problem. As such, our solution offers a robust and effective tool for risk detection in irregular time series datasets.}


\maketitle

\blfootnotetext{

\small 
\faEnvelope~rodis@uom.edu.gr,~~
\href{https://pantelisrodis.blogspot.com/}{\faLink}

\indent
\small University of Macedonia, Greece
\hfill May 2026
}


\section{Introduction}
Traditional approaches to time series classification and anomaly detection in time series data, often exhibit limitations when confronted with sparse and irregular data \cite{khan2014one}. Machine Learning (ML) methods typically require regularly sampled and dense datasets \cite{lim2023tsgm}, and thus struggle to capture long-range temporal dependencies and handle missing or unevenly spaced observations. The emergence of deep learning architectures has shown significant promise in processing sequential data due to their ability to model complex dependencies regardless of their position in the sequence, especially when it comes to transformers \cite{islam2024comprehensive, ahmed2023transformers}. However, directly applying standard transformer models to sparse and irregular time series is still a challenge.

In this work we address these limitations by introducing a framework specifically designed for sparse and irregular time series classification. Our proposed approach leverages a transformer autoencoder augmented with a local attention mechanism. This architecture is carefully crafted to exploit the powerful pattern identification capabilities of transformers while simultaneously integrating traditional data cleaning and normalization techniques to enhance data quality.

The transformer's attention mechanism is an active field of research, primarily focusing on reducing its computational complexity \cite{zhang2024ecg} and enabling a more refined and effective investigation of relationships among input data elements \cite{xiang2024unidirectional, huang2022masked}. In this context, we investigate its use in handling irregular time series.

To demonstrate the efficacy of our framework, we apply it to a real-world case study regarding the risk estimation of non-technical losses in electrical power systems across a wide area of Greece. The identification of these losses represents a critical challenge for power utilities worldwide, as they result in substantial economic drains and operational inefficiencies \cite{depuru2011electricity}. Furthermore, this issue is particularly pronounced in Greece, posing a major concern for local power systems \cite{papadimitriou2017non}, where ML approaches have been explored in the past to address this problem \cite{perifanis2024towards}. The intricate nature of identifying non-technical losses is further complicated by the inherent sparse and irregular characteristics of the available data. Through extensive experimentation, we showcase that our method effectively extracts highly discriminative latent
features, leading to effective classification of suspicious cases.

The core contribution of this work is a framework designed to effectively handle irregular and sparse time series. This framework uniquely combines classical data cleaning and normalization methods with a state-of-the-art transformer autoencoder architecture, offering several key features:

\begin{itemize}
    \item It circumvents the need for interpolation of missing data {by efficiently recognizing patterns within existing sparse data}.
    \item It implements a two-step normalization procedure specifically engineered to address the challenges of data sparsity and heterogeneity.
    \item Effectively leverages local attention within the transformer autoencoder architecture for handling sparse time series and exhibits its advantages; a topic with limited references in literature.
    \item Demonstrates a robust ability to perform classification, achieving strong results {detached from the need for labeled data. Furthermore, we investigate the conditions under which training on a subset of the available dataset} is sufficient to classify the entire set.
    
\end{itemize}

The application of the framework on the risk analysis of non-technical losses, contributes a novel methodology designed to overcome inherent limitations often encountered during data collection in these systems.

The rest of the paper is organized as follows. In the next section, we present the architecture, the functionality and the training procedure of the transformer autoencoder encompassed in our framework. In section~\ref{sec:sparsityirregularity}, we discuss how we handle sparsity and irregularity issues within the time series dataset. Subsequently, in section~\ref{sec:casestudy}, we present our solution on risk estimation on non-technical losses in electrical power systems using our framework. In section~\ref{sec:evaluation}, we evaluate the performance and efficiency of the framework on the case study by comparing it with other state-of-the-art and popular methods. Related works and our conclusions are discussed in sections~\ref{sec:related} and \ref{sec:conclusions} respectively.

\section{Transformer Autoencoders for Time Series Classification}
\label{sec:autoencoders}

\subsection{Architecture}

\begin{figure}[htbp]
  \centering
  \includegraphics[width=1.0\columnwidth]{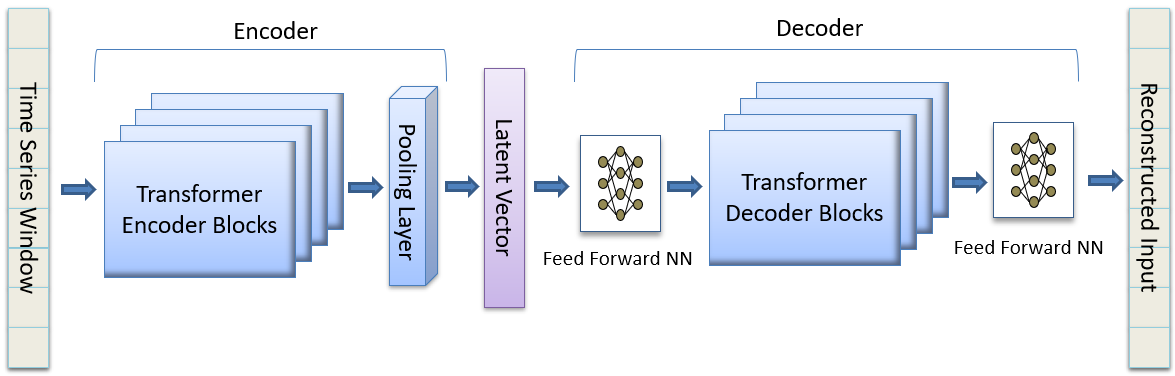}
  \caption{{Transformer autoencoder.}}
  \label{fig:transformer-aec}
\end{figure}

Transformer autoencoder is a neural network (NN) architecture comprising three key components an \textit{encoder}, a \textit{decoder} and an intermediate \textit{latent space}, \textit{i.e} a hidden vector \cite{choi2020encoding}. {An overview of the architecture we use in this work, is presented in fig.~\ref{fig:transformer-aec}}.

The core concept behind its operation is the ability to learn efficiently low-dimensional representations of input data. 
Specifically, the encoder maps the high-dimensional input sequences dataset $X$ to lower-dimensional representations $Y$ within the latent space and generates a representation model for the whole dataset of sequences. As such the encoder defines function $Y=enc(X)$.

Subsequently, the decoder attempts to reconstruct the sequences solely from their latent space representations $Y$ defined in function $X'=dec(Y)$. 
The magnitude of the reconstruction errors, computed as the Mean Square Error {$MSE(X, X')~=~1/\vert X \vert~\|X - 'X\|^2$}, indicates the extent of similarity among the sequences. 
High reconstruction errors typically suggest that input sequences deviate significantly from the patterns learned by the autoencoder during training and indicate potential anomalies or structural differences, while low reconstruction errors typically indicate common patterns among the sequences.
{We denote as $RE_A(X, N)$ the average reconstruction error of set $N$ when fed into autoencoder $A$ trained on dataset $X$.}

\begin{figure}[htbp]
  \centering
  \includegraphics[width=0.7\columnwidth]{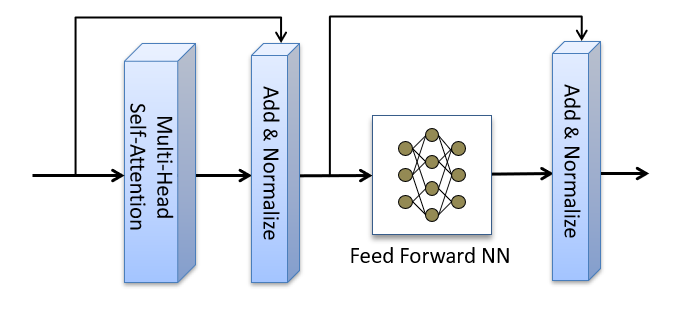}
  \caption{{Typical transformer encoder block.}}
  \label{fig:encoder}
\end{figure}

In particular, the transformer encoder architecture in our framework takes as input a time series dataset.
The encoder consists of a sequence of local attention transformer encoder blocks. Each block, the structure of which is sketched in fig.~\ref{fig:encoder}, encompasses a Multi-Head Self-Attention mechanism \cite{beltagy2020longformer} which performs several independent self-attention calculations in parallel.

The self-attention mechanism in transformers calculates attention scores between every pair of elements in a sequence, as such
allows the model to weigh the importance of different parts of an input sequence. It was introduced in \cite{vaswani2017attention} and is described as follows

\[
    \text{Attention}(Q, K, V) = \text{softmax}\left(\frac{QK^T}{\sqrt{d_k}}\right)V
\]

\begin{itemize}
    \item $Q$: query matrix
    \item $K$: key matrix
    \item $V$: value matrix
    \item $d_k$: dimension of the key vectors
\end{itemize}
The attention function is computed in a set of queries stored in matrix $Q$ and on the keys and values found into matrices $K, V$.
This design can be computationally expensive and in many time series applications this can be ineffective as the most significant dependencies or anomalous patterns might occur within a localized temporal context.

Local attention addresses these challenges by restricting the attention computation to a predefined local attention window around each element \cite{ramachandran2019stand}. This reduces the computational complexity and guides the model to primarily focus on immediate temporal neighbors, which is often beneficial for learning fine-grained patterns and changes within time series data. This might look similar to the sliding window technique \cite{provotar2019unsupervised}, the difference is that local attention aims to apply the attention mechanism on parts of the time series, while the sliding window splits the training sequence into smaller segments to feed them into autoencoder.

Essentially, during local attention an element still generates its query vector $q_i$ from the query matrix Q. However, it only attends to key-value pairs $(k_j, v_j)$ where index $j$ falls within the local window of size $w$ around element $i$.  This is typically implemented by applying a mask to the attention scores $QK^T$ such that values outside the window are ignored during the softmax operation.

{
The masking procedure handles the time series edges by adjusting the window size for elements near the boundaries. For any element $i$ of time series $x$, the window's span is computed so as not to exceed the time series length.
The start index of the window is set to 
$\max (0,i- \lfloor w/2 \rfloor)$
while the end of the window is set to 
$\min (\vert x \vert,i+ \lfloor w/2 \rfloor +1)$.
This approach ensures that elements at the boundaries fall within a smaller but valid attention window.
}

{
The resulting mask is then passed to the multi-head attention layer within the transformer encoder block. The attention mechanism uses this mask to zero out the attention scores for any elements outside the local window, effectively preventing them from influencing the output during the softmax operation. This ensures that the model focuses exclusively on immediate temporal neighbors.
}

The outcome of the final transformer encoder block is passed to a Global Average Pooling layer applied across the sequential dimension. The resulting pooled vector is then passed to a final Dense layer to project it into the $n$-dimensional vector of the latent space. This vector serves as the compressed, meaningful representation of the input time series window in $n$ dimensions.

\begin{figure}[htbp]
  \centering
  \includegraphics[width=0.7\columnwidth]{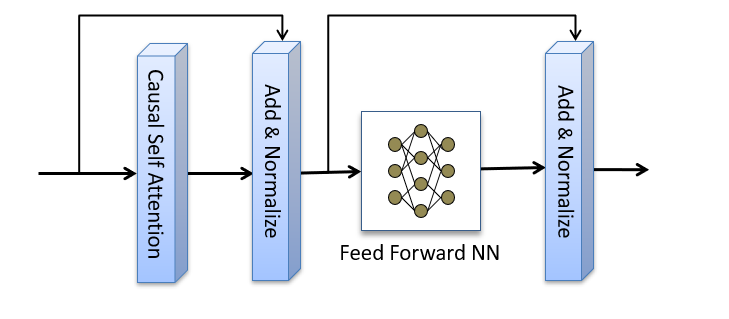}
  \caption{{Typical transformer decoder block.}}
  \label{fig:decoder}
\end{figure}

{Subsequently, the decoder strives to reconstruct the original input time series from their latent space representation. A typical decoder is sketched in fig.~\ref{fig:decoder}. The decoder first passes the latent representation through a dense layer and a reshape operation to match the required time series sequence shape.}

{Then the reshaped representations are fed to a sequence of Transformer Decoder Blocks. Each block incorporates a Causal Self-Attention mechanism, followed by a Feed-Forward Network. Commonly but not always necessary, layer normalizations are applied throughout the block to ensure stable training. A final dense layer projects the output back to the original feature dimension to produce the reconstructed time series.}

\subsection{Training and Inference}

{As a neural network architecture, autoencoders are trained on a dataset $D_s$ and then they can be used for inference on dataset $D_w$ which may be identical to $D_s$ or a distinct evaluation set. The training objective is to minimize the reconstruction error which is commonly defined as the Mean Square Error between the original and the reconstructed samples. During inference, the autoencoder produces the latent space representation, the reconstructed sample and the associated reconstruction error calculation.}

{In this work we focus on cases where $D_s \subseteq D_w$ and we aim to perform unsupervised clustering of a large dataset by training an autoencoder on a subset of it. The goal is to identify and group time series in $D_w$ exploiting their structural similarities with the time series in $D_s$, revealed in the learned latent space. As such, a cluster containing low reconstruction errors of the original training samples is expected during inference to contain samples of low reconstruction errors from $D_w$. The latent space generated from $D_s$ will exhibit approximately the same discriminative power and intrinsic dimensionality as a latent space generated from training the autoencoder on $D_w$.}

{The training results can be used for clustering by examining two primary characteristics of the autoencoder:}

\begin{itemize}
\item {The quality of the latent representation. In this design, clustering methods like the K-Means algorithm can be applied to cluster the dataset $D_w$ samples into similarity groups based on their generated latent space representations.}

\item {The reconstruction error, which can be used for grouping elements of $D_w$ by their projection distance from the data manifold learned by the autoencoder. Outliers and anomalies will yield a high reconstruction error, while samples structurally similar to the majority of samples in $D_s$ will yield low reconstruction errors. This allows for a threshold-based classification for the samples in $D_w$.}
\end{itemize}

We can use $D_s \subset D_w$ for training with the same results as using $D_w$ provided that $D_s$ is representative of the entire dataset. Let us define the notion of representative subset:

\begin{definition}[Representative Subset]
\label{definition1}
{For autoencoder $A$ and sets $D_w$ and $D_s~\subset~D_w$. Set $D_s$ is representative of $D_w$ if for acceptable threshold $t$, it holds that 
$RE_A(D_w, D_w)~-~RE_A(D_s, D_w)<t$.} 
\end{definition}

{
The value of threshold $t$ is always case-dependent and relevant to the objectives of each problem. Therefore, it is necessary to determine whether training the autoencoder with sets $D_w$ and $D_s$ produces acceptable and similar results and subsequently assess the value of $t$ as acceptable.
}

{Conversely, if the training set lacks representativeness, 
\textit{e.g.} $D_w$ contains time series that do not exhibit similarities with elements of $D_s$, the two learning paradigms will produce divergent results.}  

\subsection{Clustering and evaluation}
\label{sec:kmeans}

{In this work we consider reconstruction error-based clustering, which provides a straightforward method for grouping samples into groups of normal, low error, samples and groups of high error outliers. It is always a problem-dependent case to determine which category we are most interested in.}

{During inference, reconstruction errors are recorded for all time series in $D_w$. The dataset is typically sorted in ascending order if we care for normal time series and in descending when we care for outliers. 
Then we either form clusters of fixed size and split the sorted dataset in them or we use threshold criterion $c$ to assign each time series to a cluster comparing reconstruction errors with $c$.} A challenging issue is determining the optimal cluster's fixed size or the value of $c$, as it requires foreseeing the distribution of common features in the time series dataset. This is a problem-dependent feature and as such there is not a standard procedure to determine it. 

{To evaluate the clustering outcome we may use labeled samples in $D_w$ and examine their distribution in the generated clusters. In this design, labeled data are only used for evaluation and do not alter the unsupervised nature of the training procedure.}
The most successful clustering, manages to place (most of) the labeled data in the same cluster and maintains the size of the cluster in some acceptable range. As such, we choose cluster $C_k$ of size within acceptable range which maximizes recall {and precision} as defined next.

\begin{itemize}
    \item \textbf{Recall or True Positive Rate}. For cluster $C_k$ we compute 
    \[Recall=\frac{TP}{(TP+FN)}\]
    where True Positive $(TP)$ are the labeled data within $C_k$ and False Negative $(FN)$ are the labeled time series not classified in this cluster.  This metric assesses how comprehensively $C_k$ captures the members of a known labeled set.
    \item {\textbf{Precision.} Quantifies the purity of a cluster with respect to the known positive class.
    \[Precision=\frac{TP}{\vert C_k \vert }\]
    Note that in case of a small labeled subset within a large cluster $C_k$, the precision is expected to be numerically small.}
\end{itemize}

In the context of risk estimation, and given that $D_w$ encompasses a number of positive labeled samples, the cluster containing the largest number of positive samples includes time series of similar structure to these positive samples, as such it forms the higher-risk class. Conversely, clusters with fewer positive samples indicate lower-risk classes.

\section{Irregularity and Normalization}
\label{sec:sparsityirregularity}

The method presented in this study primarily addresses the challenging issue of detecting patterns in time series which exhibit sampling sparsity and irregularity in data collection. Sparsity occurs when the average sampling rate is too low for traditional pattern recognition methods to be effective. 

Irregularity can arise when data for the same features are gathered from diverse, heterogeneous sources with differing specifications and varied scaling. For instance, in the case study presented in section~\ref{sec:casestudy}, power consumption measurements were collected by humans and two distinct types of telemeters. Each source follows unique procedures and operates within varying time intervals. Additionally, these measurements represent accounts from diverse consumption profiles \textit{e.g.} houses, factories, public institutions and cause variation in the scaling of the recorded data.

Crucially, sparsity combined with variations in sampling rate creates time series where relationships between features are loose. {These characteristics impose the use of unsupervised learning rather than depending on the sparse labeled data for some supervised alternative. Furthermore, the inherent diverse scaling necessitates pre-processing of the raw data and normalization before training so as to ensure that the autoencoder learns structural patterns rather than features tied to specific collection sources or magnitude. The following two subsections describe how we handle normalization issues within our framework.}

\subsection{Two-Step Normalization}
\label{sec:normalization}

In the time series dataset we apply a two-step normalization procedure, first a row-wise normalization and then the application of MinMax Scaler.
\newline

\textbf{Row-wise normalization}. This initial step transforms absolute values within each data sample, \textit{i.e.} time series, into relative proportions. The normalized value $x_{i,j}'$ on place $j$ of time series $x_i$ of $z$ time steps is computed as follows 

\[x_{i,j}'=\frac{x_{i,j}}{\sum{x_{i,z}}}\]

\noindent
This step is utilized because the distribution of values within each time series is of greater interest than their absolute magnitudes and enables the model to identify pattern variations, rather than solely anomalies based on merely extreme absolute values. This prevents the autoencoder from learning to reconstruct high-magnitude samples at the expense of those with lower magnitudes. The outcome of this normalization step highlights the proportional relationships of the values within each time series. In our case study, given that the dataset represents varying consumption sums across different accounts, row-wise normalization effectively equalize these disparities.
\newline

\textbf{MinMax Scaler}. It is a scaling technique which transforms the range of independent variables to a specified interval, in our case [0,1]. This process ensures that all features are on a uniform scale, which is crucial for algorithms sensitive to input magnitudes, such as neural networks, and is computed as follows 

\[x_{i,j}' = \frac{x_{i,j} - x_{\text{min}(j)}}{x_{\text{max}(j)} - x_{\text{min}(j)}}\]

\noindent
where $x_{i,j}'$ is the normalized value of the element in the $i$-th row and $j$-th column, $x_{\text{min}(j)}$ and $x_{\text{max}(j)}$ represent the minimum and maximum values of the $j$-th column across the dataset.
Even after applying row-wise normalization the individual values within each time series can still vary significantly across the entire dataset. MinMaxScaler will scale these proportions globally across all rows and columns to the desired range ensuring all input features to the autoencoder are within a consistent range, which is vital for the stable training and optimal performance of autoencoders and neural networks in general.

\subsection{Normalization Layers}

{The typical design of a transformer autoencoder encompasses normalizations layers as sketched in fig.~\ref{fig:encoder} and ~\ref{fig:decoder}. They are utilized to normalize the feature vector within a single sample and across all feature dimensions so as to ensures that the inputs to the subsequent dense layer, or the next transformer block, has a consistent scale and mean.}

{The application of the two-step normalization process in certain cases may render the use of normalization layers within the autoencoder architecture redundant. Furthermore, over-normalization may have a negative effect on the classification procedure.}

{Consequently, the effect of the normalization layers in the learning process has to be evaluated. In cases like the study presented in section~\ref{sec:casestudy}, these layers have to be omitted. This behavior is further explained by the following considerations.}

{Row-wise normalization already standardizes the magnitude of each sample, ensuring that all time series lie within a compact and comparable numerical range.
Applying additional per-sample normalization removes meaningful amplitude information and alters the relative feature relationships within each time series.}

{The MinMax Scaler is performed before the data are fed to the model, scaling the features. The use of normalization layers within the transformer blocks further contributed to the over-normalization of the dataset affecting training and inference.}

{Given these considerations, the removal of normalization layers must be examined from both the encoder and the decoder blocks. This modification, when it is necessary to be applied, yields faster convergence and improves reconstruction fidelity.}

\section{Case Study: Risk Estimation for Non-Technical Losses in Electrical Power Systems in Greece}
\label{sec:casestudy}

Non-technical losses in Greece constituted up to 4.5\% of total power consumption between 2019 and 2023 \cite{energypress2022}. In this context, our method is applied to risk estimation primarily focusing on the classification and identification of accounts suspected of exhibiting non-technical losses. We use the dataset provided by
Hellenic Electricity Distribution Network Operator S.A. (HEDNO)
during the 2023 Datathon \cite{HEDNO2023}.

{
This dataset was chosen for its unique combination of characteristics, which provide a significant pattern recognition challenge that sets it apart from other available datasets. It is a large-scale dataset characterized by substantial heterogeneity, data sparsity and irregular sampling due to inconsistent data collection practices.
Additionally, the data records show scalar variations stemming from diverse power consumption profiles. These characteristics make the dataset an ideal case study for developing robust methods for pattern identification and risk estimation.}

{Our review of related works shows that most research on non-technical losses relies on small to medium sized datasets with frequent sampling, which do not address the diverse inconsistencies found in this dataset and the challenges associated with them. The datasets used in literature can be effectively analyzed using less sophisticated methods than the one proposed in this work.
}

\subsection{Dataset Description}
\label{sec:dataset}

The dataset contains real-world data collected from a broad geographical area across Greece including information on electricity supply companies, consumption measurements, declared consumption types per account and other additional information.
In our analysis, we hold the assumption that accounts with non-technical losses exhibit common patterns in their power consumption. 

\begin{figure}[htbp]
  \centering
  \includegraphics[width=\columnwidth]{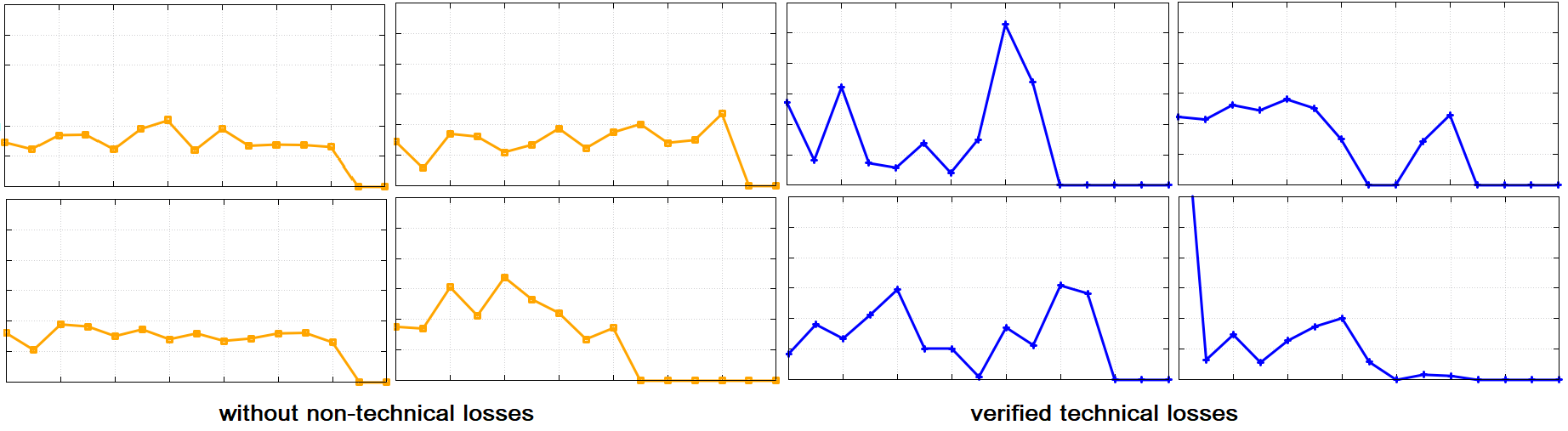}
  \caption{Indicative examples of average monthly consumption diagrams, after applying row-wise normalization, of accounts with and without verified non-technical losses.
  }
  \label{fig:consumptions}

\end{figure}

Fig.~\ref{fig:consumptions} presents indicative examples of normalized average monthly consumption diagrams for accounts with verified non-technical losses and accounts not found to participate in losses from the given dataset. These examples demonstrate that non-technical losses accounts show consumption fluctuations, while accounts without losses maintain a smoother consumption profile and this observation is also found in other studies like in~\cite{lepolesa2022electricity}. 
Fluctuations arise from interventions in the power system which lead to anomalies in consumption profiles. {Therefore we utilize the transformer autoencoder for anomaly detection, specifically seeking for outliers.}
Our analysis focuses on account-level consumption rather than household-level consumption, as residents changes in a house can introduce consumption variability and fluctuations independent of non-technical losses.

Based on these, we restrict our analysis to the dataset of consumption measurements; provided in kilowatt-hours. This dataset includes anonymized data of 1,234,509 consumer accounts and 9,201,395 consumption measurements located in an urban area in Greece collected from early 2018 to late 2022. Non-technical losses were verified for 2,316 of the accounts. The absence of explicit negative labels in the dataset, \textit{i.e.} accounts confirmed not to exhibit non-technical losses, is a reasonable and inherent characteristic of this problem domain which {favors the application of unsupervised methodologies.}

The dataset exhibits significant heterogeneity and data sparsity across several dimensions:

\begin{itemize}
\item Data collection methodologies vary, comprising manual readings by employees of HEDNO and inputs from two distinct types of telemeters one for medium voltage and another for low voltage supplies. 
\item Temporal heterogeneity in consumption periods. The accounts exhibit diverse periods of activation and overall consumption durations. Consequently, this dataset generates time series of variable lengths, a large portion of which are too short to be useful in a risk estimation model.
\item The dataset includes disparate energy consumption profiles as it comprises industrial load profiles, consumption data from small and medium-sized enterprises and residential consumption profiles which constitute the majority of the profiles.
\item The consumption data demonstrate sparsity due to their irregular measurement intervals; regardless of their consumption profile. For most accounts consumption data are recorded quarterly while a subset is measured monthly and some accounts feature arbitrary measurement periods.
\end{itemize}

\begin{figure}[htbp]
    \centering

          \centering
  \includegraphics[width=0.7\columnwidth]{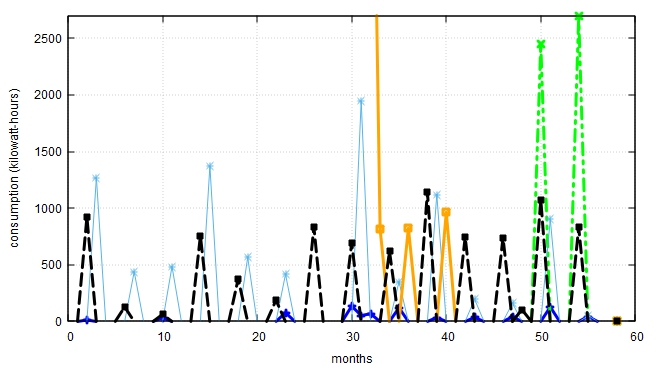}
  \caption{Consumption measurements.
  }
  \label{fig:irregs}
\end{figure}

This inherent heterogeneity, a sample of which is illustrated in fig.~\ref{fig:irregs}, mandates the transformation of the dataset into a more suitable form and the removal of invalid or noisy data. The consumption measurements have a temporal dimension and the most obvious choice is to form time series in a normalized form as described next.

\begin{figure}[htbp]
  \centering
    \centering
  \includegraphics[width=0.7\columnwidth]{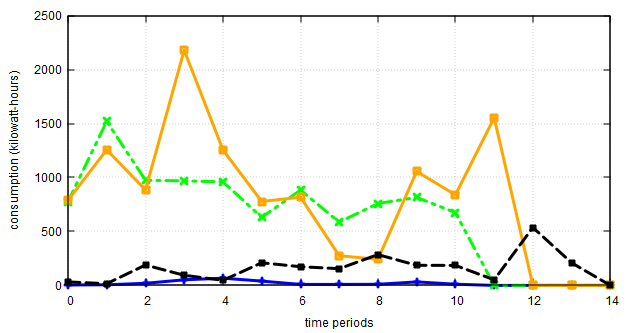}
  \caption{Average monthly consumptions before normalization.
  }
  \label{fig:unormalized}

    \label{fig:examples}
\end{figure}

\subsection{Pre-Processing and Cleaning}
\label{sec:preprocessing}

Pre-processing transforms the given dataset into a set of time series. During the cleaning procedure we remove invalid values and time series that are empty or too short to be useful in our analysis. The sequence of procedures is executed as follows:

\begin{enumerate}

\item The given dataset is transformed into a matrix, where each row corresponds to an individual account and each column represents a specific month within the time interval spanning the years 2018 to 2022. The matrix elements denote the measured consumption for the respective account in the corresponding month. Missing measurements in a given month are represented by negative values. During this transformation, invalid outlier values such as arbitrarily large readings were removed from the dataset.

\item Subsequently, labeling was applied in 2,316 accounts for which non-technical losses were confirmed.

\item The dataset is further transformed into a set of arrays which form the time series.
In every row of the matrix, for every time interval defined between two consecutive measurements we compute the average monthly consumption as the fraction of the latest measurement divided by the number of months in the interval and place the quotient into the corresponding array. The heterogeneity of the data implies that the generated time series vary in scale; a few examples are illustrated in fig.~\ref{fig:unormalized}.

\item Subsequently we apply the row-wise normalization, as discussed in detail in section~\ref{sec:normalization}, which addresses the scaling variation problem. Indicative examples of some resulted time series are illustrated in fig.~\ref{fig:consumptions}.

\item Further cleaning is necessary as time series of zero values introduce noise to the dataset and time series of length less than seven time steps cannot be efficiently exploited, they are too short to represent delinquent patterns as we concluded through extensive experimentation, therefore they are removed. 

\item To ensure consistent model analysis and considering the variable lengths of our time series, all time series are pruned and zero-padded to a uniform length of {58 time steps}.

\item Before feeding the time series to the autoencoder we apply the MinMax Scaler normalization procedure, also described in section~\ref{sec:normalization}.
\end{enumerate}

The produced dataset consists of 347,546 time series {each of 58 time steps} in normalized form and 891 of them are labeled as participating in non-technical losses activities. Each time series constitutes of a sequence of average monthly consumptions, one per measurement period. Based on our assumption and for an account that experiences non-technical losses, its behavior as reflected on its power consumption measurements will change and we will detect this by studying the pattern of the average monthly consumptions among the measurement periods.

\subsection{Risk estimation}

{Following inference, time series are sorted in descending order by reconstruction error to facilitate outlier detection. Time series of higher reconstruction errors are designated as outliers that exhibit higher risk of participating in power theft. Then we employ the strategy of dividing the time series in clusters of fixed size, where the cluster's size is set equal 15\% of $D_w$; the final cluster consists approximately of 10\% of the total. Consequently, time series classified into the first cluster exhibit the highest risk, the estimated risk is decreasing in each subsequent cluster.}

{The operation of the autoencoder is unsupervised. Labeled data are not used during training but they are used for the evaluation of the training results. In this spirit, the cluster size is chosen to harmonize the requirement for a sufficiently large percentage of labeled positive samples in the outlier cluster which will indicate high risk, with the goal of forming clusters with a number of samples that could be practically utilized, \textit{i.e.} subsequent manual investigation.}

{Our benchmarks showed that the proposed autoencoder training on either the full dataset $D_w$ or on a representative subset $D_s$ yields similar results; this is extensively analyzed in section~\ref{sec:evaluation}. As such we can avoid the computationally demanding training on $D_w$ by utilizing a smaller training dataset of 100,000 randomly chosen time series of $D_w$. Furthermore, we excluded the labeled time series from all autoencoder training configurations to avoid potential label bias and any dependency on labeled data during training.}

\section{Evaluation}
\label{sec:evaluation}

{For the evaluation of our proposed method, we utilize comparison with other popular and state-of-the-art methods and with alternative configurations. The comparison is based on two pillars. Following inference, firstly we study the distribution of the positive labeled data in the formatted clusters, quantifying the model's utility for risk estimation. 
The comparison is conducted in terms of Recall and Precision for the high-risk cluster of each configuration. Secondly, we compare the robustness of each configuration evaluating the consistency of the produced clustering across the clusters of all different methods and configurations.}

\subsection{Comparison Methods}
\label{sec:methods}

{The proposed transformer autoencoder (TR-LA) encompasses the following features
\begin{itemize}
    \item 4 transformer encoder blocks and 4 decoder blocks.
    \item Dimension of the Feed Forward NN in each transformer's block is set to 16.
    \item Every multi-head attention mechanism encompasses 5 heads, each with a dimension
    of 5.
    \item Latent space dimension is set to 10.
    \item Local attention window is set to 5.
\end{itemize}
}

\begin{figure}[htbp]
  \centering
  \includegraphics[width=0.7\columnwidth]{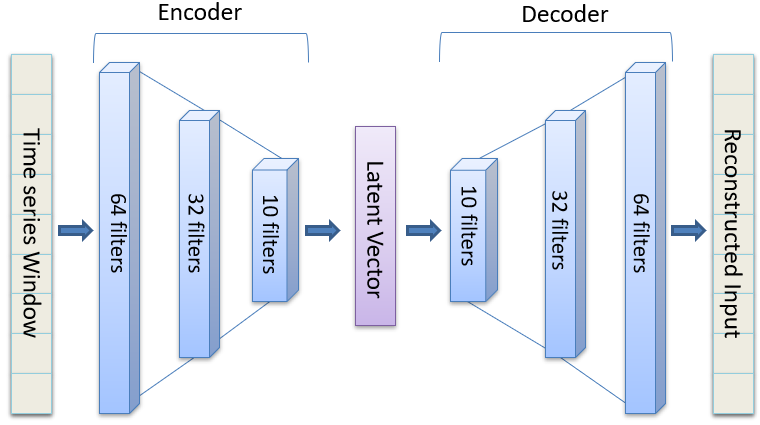}
  \caption{{Convolutional Autoencoder.}}
  \label{fig:conv-autoencoder}

\end{figure}

\noindent
{
TR-LA is compared against the following methods, all implemented using standard functionalities within the TensorFlow and Keras libraries. The hyperparameters were determined through extensive experimentation.
\begin{itemize}    
    \item Transformer autoencoder with full self-attention mechanism (TR-FU); local attention window is set equal to time series size. This comparison exhibits the advantages in local pattern recognition of the local attention mechanism.
    \item Convolutional 1D autoencoder (CONV), a standard state-of-the-art approach in this domain \cite{chen2020one}. Its architecture is sketched in fig.~\ref{fig:conv-autoencoder} and consists of the following components:
    \begin{itemize}
            \item The encoder is composed of three convolutional layers with 64, 32 and 10 filters, respectively.
            \item The decoder mirrors this structure in reverse, consisting of three layers with 10, 32 and 64 filters.
            \item The kernel size is set to 3 for all layers.     
    \end{itemize}
    \item Standard Feed Forward Deep Neural Network (FF-NN) employed for binary classification, which is a classical and popular approach~\cite{lepolesa2022electricity}. The model consists of:
     \begin{itemize}
        \item An input layer with a size equal to the time series size.
        \item Two hidden layers, each with 64 nodes utilizing the hyperbolic tangent (tanh) activation function.
        \item A single-node output layer with a sigmoid activation function to produce the final classification.
    \end{itemize}
    The NN model is trained on a fixed size dataset of size ten times the total labeled data; where verified losses are labeled as 1 and the remaining data labeled as 0. The FF-NN outputs a prediction of the probability of each time series to belong to the positive class. The resulting time series are sorted based on this predicted probability and then divided into clusters. For the evaluation of the model we employed three training configurations: FF-NN-05, FF-NN-08 and FF-NN-10 which encompass respectively 50\%, 80\% and 100\% of the labeled samples in their training sets. The models in all configurations were trained for 80 epochs.
\end{itemize}
}

{
Given the unsupervised nature of the autoencoders, it is necessary to investigate the effect of different dataset sizes in training. Therefore, we employed three training dataset configurations utilizing different subsets of the available training dataset, distinguished by different suffices on the models' notations. In this context, it is found that each configuration requires a varying number of epochs to achieve optimal training and prevent under- or overfitting, presented in table~\ref{tab:trainingsets}, and this holds both for the transformer and the convolutional autoencoders. 
Note that notation -346K indicates models trained on the entire available dataset, excluding the labeled samples. The batch size for all autoencoder models is set equal to 10,000.
}

\begin{table}[htbp]
  \centering
  \caption{{Training sizes and epochs for autoencoder training}}
  \label{tab:trainingsets}
  \begin{tabular}
  {ccc}
    \toprule
    Notation Suffix & Epochs & Training Set Size \\
    \midrule
    -346K & 20 & 346,567 \\
    -100K & 30 & 100,000  \\
    -10K  & 80 & 10,000  \\
    \bottomrule
  \end{tabular}
\end{table}

\subsection{Evaluation Results}
\label{sec:results}

{The following figures \ref{fig:TRANSF-LA}, \ref{fig:TRANSF-FU}, \ref{fig:CONV} and \ref{fig:FF-NN-10} illustrate the distribution of positive labeled samples in the formatted clusters of models TR-LA-100K, TR-FU-100K,  CONV-100K and FF-NN-10, respectively. These are the most efficient training dataset configurations based on our benchmarks.}

\begin{figure}[htbp]
    \centering
    \begin{minipage}[b]{0.47\textwidth} 
        \centering
        \includegraphics[width=\textwidth]{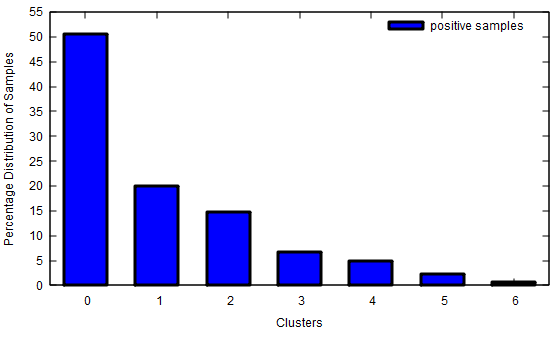}
        \caption{{Transformer autoencoder with local attention (TR-LA-100K).}}
        \label{fig:TRANSF-LA}
    \end{minipage}
    \hfill 
    \begin{minipage}[b]{0.49\textwidth} 
        \centering
        \includegraphics[width=\textwidth]{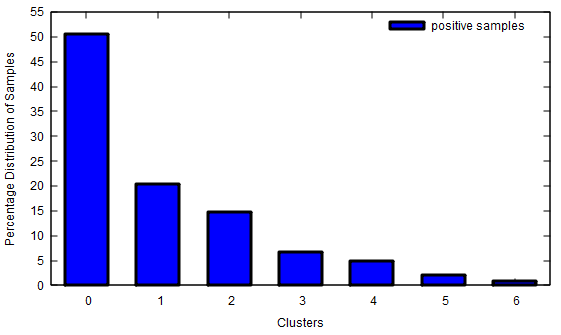}
        \caption{{Transformer autoencoder with full self-attention (TR-FU-100K).}}
        \label{fig:TRANSF-FU}
    \end{minipage}
\end{figure}

\begin{figure}[htbp]
    \centering
    \begin{minipage}[b]{0.48\textwidth} 
        \centering
          \includegraphics[width=\columnwidth]{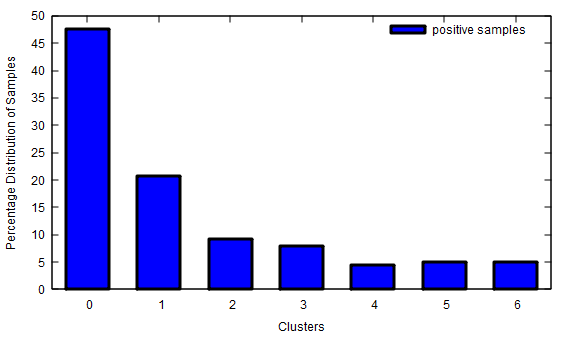}
            \caption{{Convolutional autoencoder \\ (CONV-100K).}}
            \label{fig:CONV}
    \end{minipage}
    \hfill 
    \begin{minipage}[b]{0.49\textwidth} 
        \centering
  \includegraphics[width=\columnwidth]{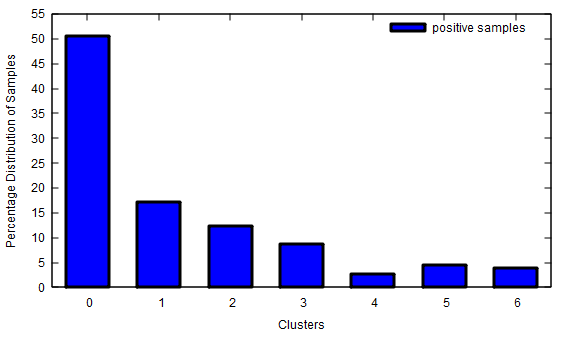}
  \caption{{Feed Forward Neural Network  \\ (FF-NN-10).}}
  \label{fig:FF-NN-10}
    \end{minipage}
\end{figure}

{The choice to use -100K subset for training is supported by the average reconstruction errors shown in Table~\ref{tab:RE} and our experimental results. As defined in Definition~\ref{definition1}, -100K serves as a representative subset for the transformer autoencoders as it yields similar reconstruction error with training in the entire dataset. Based on our benchmarks, CONV also yields similar results when trained in -100K and -346K.}

\begin{table}[htbp]
  \centering
  \caption{{Average reconstruction errors}}
  \label{tab:RE}
  \begin{tabular}{c|cccc}
    \toprule
     & -346K & -100K & -10K & \\
    \midrule
    TR-LA  & 0.000575 & 0.000581 & 0.000585 & \\
    TR-FU & 0.000525 & 0.000518 & 0.000616 & \\
    CONV & 0.000246 & 0.001072 & 0.002074 & \\
    \bottomrule
  \end{tabular}
\end{table}

{All models manage to allocate a large proportion, approximately 50\% of the labeled data, in the high-risk cluster and initially appear to approach their quantitative target. This result shows that the data pre-processing and data cleaning strategy is effective in exposing patterns detectable by all models.
As indicated by the recall and precision metrics of the high-risk cluster, on table~\ref{tab:recall}, the CONV-100K autoencoder lags slightly in efficiency, while FF-NN-10 slightly outperforms the rest. Note that due to the limited number of labeled samples, highest possible value for precision is 0.01709.}

\begin{table}[htbp]
  \centering
  \caption{{High risk cluster recall and precision}}
  \label{tab:recall}
  \begin{tabular}{c|cccc}
    \toprule
     & TR-LA-100K & TR-FU-100K & CONV-100K & FF-NN-10 \\
    \midrule
    Recall & 0.5050 & 0.5050 & 0.4769 & 0.5061 \\
    Precision & 0.00863 & 0.00863 & 0.00815 & 0.00865 \\
    \bottomrule
  \end{tabular}
\end{table}

{However, a further analysis of the supervised training requirements in FF-NN and the consistency of the results for the autoencoders reveals significant qualitative differences in the produced results, as shown next.}

{
\subsection{Robustness and Efficiency}
\label{sec:robustness}
}

{The supervised nature of the FF-NN architecture exhibits a vulnerability to the selection of different training datasets. Given the irregular data collection methods, this vulnerability can have a significant impact on the produced results. This concern is outlined in the recall metrics of all FF-NN configurations, shown in Table~\ref{tab:FFNN-recall}, and is more concretely illustrated in the subsequent consistency evaluation of all models, table~\ref{tab:concistency}. Fewer labeled samples in the training set degrade the recall of the high-risk cluster, furthermore the clustering procedure exhibits inconsistency.}

\begin{table}[htbp]
  \centering
  \caption{{Recall for FF-NN  \\ training configurations}}
  \label{tab:FFNN-recall}
  \begin{tabular}{ccc}
    \toprule
    FF-NN-05 & FF-NN-08 & FF-NN-10 \\
    \midrule
    0,4702 & 0,4713 & 0,5061 \\
    \bottomrule
  \end{tabular}
\end{table}

{The robustness of each architecture is measured in terms of clustering consistency across the different training configurations. In this respect, we measure the percentage of samples classified by each method in the same cluster across the different training dataset configurations, \textit{i.e.} -346K, -100K and -10K. 
A method that consistently assigns the same samples on the same clusters using different configurations of training datasets, exhibits robustness and successfully mitigates challenges posed by irregularities in data collection and time-series formation.}

{Table \ref{tab:concistency} presents the percentage metrics. The transformer autoencoders are more consistent, exhibiting a high percentage of consistent sample assignment to the high-risk cluster and average high consistency across all clusters, which signifies the qualitative advantages of their clustering outputs against the other methods. Specifically, TR-LA shows greater consistency than the TR-FU, highlighting the advantages of local attention and the improved discriminative abilities it provides. Conversely, FF-NN and CONV fail to provide reliable classifications as they exhibit vulnerability to variations in the training dataset formation.}

\begin{table}[htbp]
  \centering
  \caption{{Percentage of clustering similarity (\%)}}
  \label{tab:concistency}
  \begin{tabular}{ccccc}
    \toprule
    Cluster & \textbf{TR-LA} & TR-FU & CONV & FF-NN \\
    \midrule
    \#1 & \textbf{98.18} & 93.52 & 72.24 & 48.14 \\
    \midrule
    \#2 & \textbf{95.28} & 82.43 & 37.08 & 23.14 \\
    \midrule
    \#3 & \textbf{92.38} & 73.69 & 24.46 & 16.78 \\
    \midrule
    \#4 & \textbf{89.00} & 64.38 & 19.67 & 19.02 \\
    \midrule
    \#5 & \textbf{86.48} & 58.24 & 19.83 & 20.43 \\
    \midrule
    \#6 & \textbf{86.48} & 60.48 & 24.63 & 32.66 \\
    \midrule
    \#7 & \textbf{90.64} & 73.34 & 13.96 & 53.29 \\
    \toprule
    Average: & \textbf{91.21} & 72.30 & 30.27 & 30.50 \\
    \bottomrule
  \end{tabular}
\end{table}

{For the evaluation we used an NVIDIA Tesla T4 GPU equipped with 16 GB of memory and 15 GB of system RAM. Training on the entire dataset required approximately 5 minutes for the transformer autoencoders, 1 minute for CONV and 30 seconds for FF-NN. When run on a conventional desktop PC, equipped with an 8-core CPU and 8 GB of RAM, the runtime for TR-LA-346K was 2 hours, while for TR-FU-346K required 6 hours. Training the models with -100K instead of -346K, reduces runtime nearly by half for all models. These metrics highlight the computational overhead associated with transformer autoencoders and the efficiency provided by the local attention mechanism over full attention.}

{The experimental evaluation concluded in favor of the use of transformer autoencoder with local attention, demonstrating consistent classification results and reasonable computational overhead.}

\section{Related Works}
\label{sec:related}

In this section we review related works in irregular time series classification and risk estimation of non-technical losses encompassing deep learning methods.

\subsection{Irregular Time Series Classification}
\label{sec:relatedTS}

The works reviewed in this section illustrate the evolving trends in applying deep learning and particularly autoencoders and attention mechanisms to handle sparse and irregular time series.

A variational autoencoder-based model specifically designed for irregularly sampled time series is proposed in \cite{rubanova2019latent}. The autoencoder explicitly models the probability of observation times using Poisson processes, allowing it to handle arbitrary time gaps effectively. In \cite{li2020learning} the use of a variational autoencoder for irregular time series is also explored. It employs a masked input form to replace missing entries with zeros. While applied to images in their work, the underlying principle is relevant to handling missing data in time series.

A novel deep learning architecture is introduced in \cite{shukla2019interpolation} for supervised learning with sparse and irregularly sampled multivariate time series. This architecture leverages a semi-parametric interpolation network to facilitate information sharing across multiple dimensions during the interpolation stage, followed by a flexible prediction network.

The use of autoencoder with LSTM layers for unsupervised anomaly detection is discussed in \cite{provotar2019unsupervised}. This approach employs a sliding window technique to capture local temporal relationships within time series, though it was applied to a small dataset without irregularities.

In \cite{shukla2021multi} a deep learning framework incorporating Multi-Time Attention Networks is developed. It is designed for irregularly sampled time series derived from multi-source data. This framework learns continuous time embeddings and uses an attention mechanism to generate fixed-length representations from time series with varying observation counts. Experimental evaluations were carried on physiological and medical records.

The problem of Precursor-of-Anomaly Detection for Irregular Time Series is investigated in \cite{jhin2023precursor}. Their work addresses two objectives, detecting current anomalies and predicting future anomalies. They tackle this multi-task learning problem using a neural controlled differential equation-based network.

Data irregularity is addressed through the synthesis of missing data in \cite{lim2023tsgm}. A score-based generative model is proposed for this purpose, aiming to reconstruct missing observations in irregular time series.

Our work complements these efforts by introducing a new methodology that improves upon current research directions in this domain. Specifically, our framework processes existing sparse data to directly handle irregularity circumventing the need for synthesizing missing data or filling time gaps.

\subsection{Non-Technical Losses}
\label{sec:related losses}

In \cite{lin2021electricity}, an adaptive time-series recurrent neural network architecture for the detection of abnormal users is presented.
It encompasses an oversampling technique by generating a set of new samples by random linear interpolation computation on existing minority samples. The method is evaluated on a small dataset.

Synthetic Minority Over-Sampling is also used in \cite{zhu2024electricity}. The large dataset used contains daily consumption data and interpolation methods are used to address the limitations and missing data. For the detections of power theft a combination of Convolutional NN and ML techniques is used. A large dataset is also used in the evaluation of \cite{shehzad2022electricity} which encompasses daily consumptions labeled for malicious and normal users. The proposed method incorporates synthetic data generation and the use of a Genetic Algorithm on the labeled data.

NN architectures enhanced with data synthesis procedures are proposed in \cite{lepolesa2022electricity} and \cite{el2023electricity} yielding promising results. The former is applied in a medium size data set while the latter in a small dataset.

Convolutional autoencoder is used in \cite{cui2021two} for electricity theft identification.
The autoencoder extracts and identifies the abnormalities of electricity
theft users against the uniformity and periodicity of normal power
consumption features. The outcome of the autoencoder is further processed with ML techniques to build a prediction model. This study is applied in a small dataset with half an hour sampling.

Autoencoder for non-technical losses is also used in \cite{javaid2022non} which also combines synthetic data procedures. The experimental application is conducted in a medium size dataset with daily data sampling.

Using self-attention mechanisms for electricity theft detection is not a novel concept. In \cite{finardi2020electricity} an attention augmented
convolutional NN architecture is proposed which exhibits interesting results in comparison to other deep learning methods. The experimental evaluation of the method is applied on a medium size dataset of labeled data enhanced with interpolated data.

NN autoencoders combined with ML techniques for electricity theft detection are used in \cite{yan2021electricity} and \cite{huang2021electricity}, yet the methods are evaluated against small datasets drawing limited conclusions.
In similar design, a variational autoencoder equipped with an attention mechanism is proposed in \cite{takiddin2022deep}, presenting interesting results when applied in a small size dataset formed by half an hour sampling

{Probabilistic forecasting using deep learning models is studied in \cite{lu2022probabilistic}. 
It is a concept closely related to the analysis of uncertainties and risks like non-technical losses and this study provides a useful method for load analysis and forecasting in power systems, with a particular focus on residential daily electricity consumption. 
In the same spirit, \cite{tarmanini2023short} studies load forecasting on residential load profiles using machine learning techniques, specifically exhibiting the advantages of NN over traditional regression methods.}

Most of the aforementioned works are applied to small datasets, raising questions about their efficiency in big data. The datasets used for evaluation are typically formed through dense data collection and do not adequately address data sparsity, irregularities or variable sampling rates. Similar to many methods in the literature, they rely on interpolation for missing data and employ processes for synthesizing labeled samples rather than directly addressing classification challenges.

Conversely, our framework directly addresses sparsity and irregularity by leveraging existing data and tackles the challenge of missing labels, eliminating the need for interpolation or synthetic data generation. Moreover, the evaluation dataset that we used is large and exhibits the efficiency of our approach, a characteristic shared by few other works in literature.

\section{Conclusions}
\label{sec:conclusions}

{The transformer architecture is a powerful tool for time series analysis and classification. However, effectively handling sparse and irregular time series presents significant challenges that necessitate a more sophisticated approach.
Encompassing local attention enables the transformers to identify local relationships among the elements of a sequence, while data cleaning and normalization techniques enable the model to identify high-level pattern in a dataset. These two enhancements, which form the core of our framework, empower the model with sufficient discriminative abilities when working with irregular time series datasets.}

The efficacy of our proposed model is demonstrated in the context of risk estimation for non-technical losses in electrical power systems. This application highlights the model's capacity for performing efficient classification and to effectively overcome data irregularities. Given the significant importance and inherent complexity of non-technical losses risk estimation, this application clearly showcases the advantages of our method.


\bibliography{sn-bibliography}

\end{document}